\documentclass{ecai}
\usepackage{times}
\usepackage{graphicx}
\usepackage{latexsym}

\usepackage{hyperref}       
\usepackage{url}            
\usepackage{booktabs}       

\usepackage{amsmath}
\usepackage{cleveref}
\usepackage{graphicx}
\usepackage{multirow}
\usepackage{amssymb}

\usepackage{amsmath,amsfonts,bm}

\def\eqref#1{equation~\ref{#1}}

\def\1{\bm{1}}

\def\vzero{{\bm{0}}}
\def\vone{{\bm{1}}}

\def\vx{{\bm{x}}}

\def\vz{{\bm{z}}}

\def\mI{{\bm{I}}}

\DeclareMathAlphabet{\mathsfit}{\encodingdefault}{\sfdefault}{m}{sl}
\SetMathAlphabet{\mathsfit}{bold}{\encodingdefault}{\sfdefault}{bx}{n}

\def\gL{{\mathcal{L}}}

\def\gN{{\mathcal{N}}}

\def\gU{{\mathcal{U}}}

\newcommand{\E}{\mathbb{E}}

\newcommand{\R}{\mathbb{R}}

\Crefname{equation}{Eq.}{Eqns.}
\Crefname{figure}{Fig.}{Figs.}
\creflabelformat{equation}{(#2#1#3)}

\newtheorem{proposition}{Proposition}

\begin{document}

\title{Regularized Cycle Consistent Generative Adversarial Network for Anomaly Detection}

\author{Ziyi Yang\institute{Stanford University, 
USA, email: ziyi.yang@stanford.edu} \and Iman Soltani Bozchalooi\institute{Ford Greenfield Labs,
USA, email: isoltani@ford.com} \and Eric Darve\institute{Stanford University,
USA, email: darve@stanford.edu} }

\maketitle
\bibliographystyle{ecai}

\begin{abstract}
In this paper, we investigate algorithms for anomaly detection. Previous anomaly detection methods focus on modeling the distribution of non-anomalous data provided during training. However, this does not necessarily ensure the correct detection of anomalous data. We propose a new Regularized Cycle Consistent Generative Adversarial Network (RCGAN) in which deep neural networks are adversarially trained to better recognize anomalous samples. This approach is based on leveraging a penalty distribution with a new definition of the loss function and novel use of discriminator networks. It is based on a solid mathematical foundation, and proofs show that our approach has stronger guarantees for detecting anomalous examples compared to the current state-of-the-art. Experimental results on both real-world and synthetic data show that our model leads to significant and consistent improvements on previous anomaly detection benchmarks. Notably, RCGAN improves on the state-of-the-art on the KDDCUP, Arrhythmia, Thyroid, Musk and CIFAR10 datasets.
\end{abstract}
\section{INTRODUCTION}
Anomaly detection refers to the task of identifying anomalous observations that deviate from what are believed to be normal data. It has been an important and active research area in many domains, such as medical diagnosis \cite{schlegl2017unsupervised}, cyber intrusion detection \cite{buczak2016survey} and robotics \cite{park2016multimodal}. Emerging deep learning models \cite{goodfellow2014generative} with extraordinary capacity to estimate the complex distributions in high-dimensional data provide new approaches for anomaly detection. Efforts have been made to address anomaly detection by deep neural networks, for example energy-based model \cite{zhai2016deep} and deep Gaussian mixture model \cite{zong2018deep}.

Recently proposed bi-directional Generative Adversarial Networks (GANs), including Adversarial Learned Inference (ALI) \cite{dumoulin2016adversarially} and ALI with Conditional Entropy (ALICE) \cite{li2017alice}, allow for high-quality mapping back from data to latent variable space via an encoder network. Adversarially Learned Anomaly Detection (ALAD) in \cite{Zenati2018AdversariallyLA}, which is built upon ALICE, leverages reconstruction of data from GAN for anomaly detection. Despite inspiring progress in deep anomaly detection, most of the previous work focuses on density estimation based primarily on normal data, i.e. the generation of normal data. However, this does not guarantee the detection of anomalous data. For instance, methods that rely on GAN start making spurious predictions when the generator network is able to (incorrectly) generate points outside the normal manifold. A theoretical understanding on how and why generative models can detect anomaly is still lacking.

In this paper, we propose a theoretically grounded algorithm based on GAN with special modifications to the loss function and discriminator networks to bias both the generator and the discriminator towards the normal manifold. We introduce a penalty distribution $t(\vx)$ w.r.t. the normal data distribution $q(\vx)$ in the adversarial training such that the generation from the latent space is biased towards normal manifold and discriminator is trained adversarially to assign higher probability to normal data. The penalty distribution is chosen to be a random noise distribution (e.g., Gaussian or uniform distribution) and the motivation will be explained in \cref{sec:meth}. Mathematical proofs show that the introduction of the penalty distribution results in a more consistent detection of anomalous data (avoiding false positive). Results on synthetic data also demonstrate that RCGAN yields more faithful generators and discriminators for anomaly detection than previous GAN-based models.

We evaluate our approach on real-world datasets including KDDCUP (network intrusion), Arrhythmia, Thyroid (medical diagnosis), Musk (molecular chemistry) and CIFAR10 (vision). On all these datasets, RCGAN outperforms all other baseline models by significant margins.

In summary, our key contributions are two-fold:
\begin{itemize}
	\item The introduction of the penalty distribution $t(\vx)$ to GAN-based framework for anomaly detection to bias the generator and the discriminator towards the normal manifold.
	\item Mathematical proofs show that RCGAN enforces a large reconstruction error on anomalous data and encourages accurate reconstruction for normal data, providing a theoretical guarantee for reliable anomaly detection.
\end{itemize}

\section{RELATED WORK}

Also known as novelty detection and outlier detection, anomaly detection has been extensively studied in literature. Previous methods can be roughly categorized into two types, representation learning and generative model.

Representation learning methods address anomaly detection by extracting common features or learning a data mapping from normal data. One-Class Support Vector Machines (OC-SVM) \cite{scholkopf2000support} finds a maximum margin hyperplane such that mapped normal data are separated from the origin. Deep Support Vector Data Description (DSVDD) \cite{ruff2018deep} optimizes a hypershpere to enclose the network representations of the normal data. ODIN \cite{liang2017enhancing} utilizes temperature scaling and perturbations upon a pre-trained neural network for image anomaly detection. In \cite{golan2018deep} researchers develop an approach for vision anomaly detection by training a classifier on geometric-transformed normal images. The classifier essentially provides feature detectors with softmax activation statistics that can be used to compute anomaly scores.

Generative models mostly try to learn the reconstruction of data and detect anomaly through reconstruction profiles. For example, autoencoders are used to model the normal data distribution and the anomaly scores are computed as the reconstruction loss or how likely a sample can be reconstructed \cite{an2015variational,nguyen2019anomaly, pidhorskyi2018generative}. In \cite{sabokrou2018adversarially} researchers introduce distorted normal data to the training of autoencoders, however, theoretical analysis is lacking why the method works. Deep Structured Energy Based Models (DSEBM) \cite{zhai2016deep} learn an energy-based model to map each example to an energy score. Deep Autoencoding Gaussian Mixture Model (DAGMM) \cite{zong2018deep} estimates Gaussian mixture from normal data via an autoencoder network. Recently, Generative Adversarial Networks have been explored for anomaly detection. GANs are leveraged to identify disease markers on tomography images of the retina \cite{schlegl2017unsupervised}. Images are mapped back to the latent variable space by a recursive backpropagation process. ALAD \cite{Zenati2018AdversariallyLA} adopts a bi-directional GAN framework and data are projected to the latent space by the encoder network. 

Among all generative methods, our work is mostly related to ALAD. In contrast to ALAD, RCGAN utilizes samples from an \textit{a priori\/} chosen random noise distribution as adversarial data during training and enables discriminators to better recognize anomalies, as suggested by our theoretical analysis and experimental evaluations.

\section{PRELIMINARIES}

In this section we briefly introduce the anomaly detection problem from a statistical angle. Then we go through closely related GAN frameworks and their applications for anomaly detection. Finally, we motivate why the application of penalty distribution in the training of GANs is essential.

\subsection{Anomaly Detection from a Statistical Perspective}

The anomaly detection problem can be formulated as follows. We consider that the ``normal'' data is defined by a probability density function $q(\vx)$ (``normal'' does not refer to Gaussian distribution in our paper unless mentioned). In unsupervised anomaly detection, during training, we only have access to samples from $q(\vx)$. The goal is to learn an anomaly score function $A(\vx)$ such that, during the \textbf{test} phase, anomalous examples are assigned with larger anomaly scores than normal examples.

\subsection{Generative Adversarial Networks}

One approach for anomaly detection is to model the underlying distribution of normal data $q(\vx)$ based on training normal examples. Generative Adversarial Networks (GANs) \cite{goodfellow2014generative} can model a distribution using a transformation $G(\vz)$ from a latent space distribution $p(\vz)$ to the space $\vx$; the \textit{generator} network $G(\vz)$ defines the conditional distribution $p(\vx|\vz)$. The generator distribution is defined as $p(\vx) = \int p(\vx|\vz)p(\vz) d\vz$.

The GANs framework trains a \textit{discriminator} network $D(\vx)$ to distinguish between real data from $q(\vx)$ and synthetic data generated from $p(\vx)$. The minmax objective function for GANs is:
\begin{equation}
\label{eq:gan}
\begin{aligned}
\min_{G}\max_{D}V(D, G) &= \E_{\vx\sim q(\vx)}[\log D(\vx)]\\
&+ \E_{\vz\sim p(\vz)}[\log(1 - D(G(\vz)))]
\end{aligned}
\end{equation}
\cite{goodfellow2014generative} shows that the optimal generator and discriminator correspond in \Cref{eq:gan} to a saddle point such that the generator distribution matches the data distribution $p(\vx) = q(\vx)$.

Adversarially Learned Inference (ALI) from \cite{dumoulin2016adversarially} introduces an \textit{encoder} network $E(\vx)$ and attempts to match the encoder joint distribution $q(\vx, \vz) = q(\vx)e(\vz|\vx)$ and the generator joint distribution $p(\vx, \vz) = p(\vz)p(\vx|\vz)$, where $e(\vz|\vx)$ is parameterized by the encoder network. The same idea is also proposed in \cite{donahue2016adversarial}. The optimization objective for ALI is defined as:
\begin{equation}
\begin{aligned}
\label{eq:ali}
\min_{E, G}\max_{D_{\vx\vz}}V_\text{ALI}(D_{\vx\vz}, G&, E) = \E_{\vx\sim q(\vx)}[\log D_{\vx\vz}(\vx, E(\vx))] \\
&+ \E_{\vz\sim p(\vz)}[\log(1 - D_{\vx\vz}(G(\vz), \vz))]
\end{aligned}
\end{equation}
where $D_{\vx\vz}$ is a discriminator network that takes both $\vx$ and $\vz$ as input and the output is the probability that $\vx$ and $\vz$ are from $q(\vx, \vz)$. It follows that the optimum of the encoder, generator and discriminator is a saddle point of \Cref{eq:ali} if and only if $q(\vx, \vz) = p(\vx, \vz)$. Also if a solution of \Cref{eq:ali} is achieved, the marginal and joint distributions in $(\vx, \vz)$ match.

In order to address the non-identifiability issues in ALI, \cite{li2017alice} proposes ALI with Conditional Entropy (ALICE) that adds a second discriminator $D_{\vx\vx}$ network to ALI to distinguish $\vx$ and its reconstruction $\hat{\vx} = G(E(\vx))$. An extra term is included in the overall optimization objective, written as:
\begin{equation}
\label{eq:alice}
\begin{aligned}
\min_{E, G} \max_{D_{\vx\vz},D_{\vx\vx}} V_\text{ALICE} &= V_\text{ALI} + \E_{\vx\sim q(\vx)}[\log D_{\vx\vx}(\vx, \vx)\\
&+ \log (1 - D_{\vx\vx}(\vx, \hat{\vx}))]
\end{aligned}
\end{equation}
\cite{li2017alice} shows that \Cref{eq:alice} approximates an upper bound of the conditional entropy $H^{\pi}(x|z) = -\E_{\pi(\vx, \vz)}[\log\pi(\vx|\vz)]$, where $\pi(\vx, \vz)$ represents the matched joint distribution $\pi(\vx, \vz) \triangleq q(\vx, \vz) = p(\vx, \vz)$. It follows that the corresponding optimal generator and encoder in \Cref{eq:alice} theoretically guarantees a perfect reconstruction for $\vx \sim q(\vx)$.

Recently, efforts have been made to utilize GANs for anomaly detection, especially bi-directional GANs mentioned above that can readily reconstruct a data example via the encoder and generator network. For example, Adversarially Learned Anomaly Detection (ALAD) in \cite{Zenati2018AdversariallyLA} trains ALICE model on normal data, where they add an extra discriminator $D_{\vz\vz}$ to encourage cycle consistency in the latent space. The anomaly score $A(\vx)$ depends on how well $\vx$ can be reconstructed.

\begin{figure}
\centering
\includegraphics[width=0.95\columnwidth]{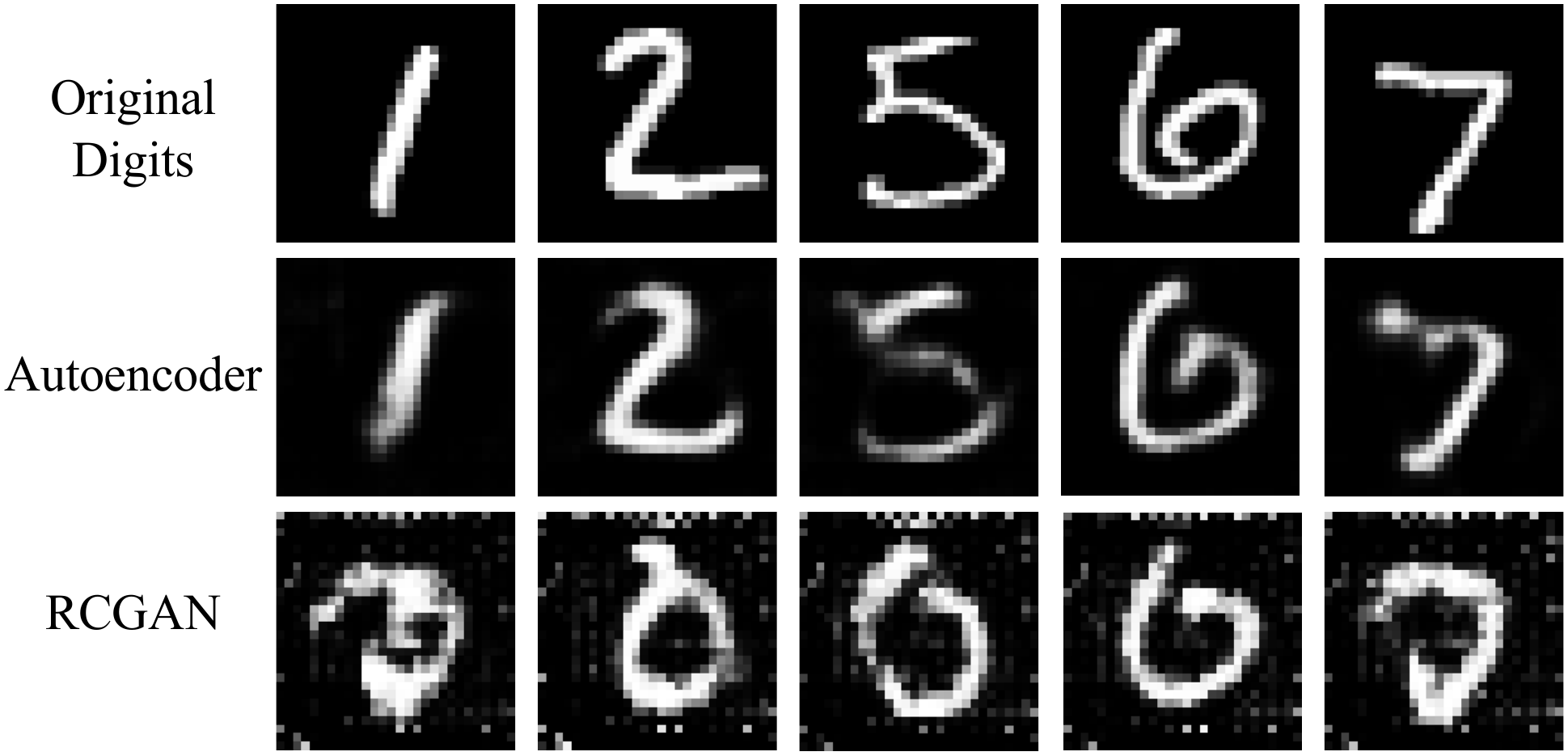}
\caption{Examples of reconstructed abnormal data by RCGAN and an autoencoder (AE), which both are trained on the normal images of digit 0. The first row includes five examples of original abnormal digits. The second and third row contain reconstructed images by AE and RCGAN respectively. AE (wrongly) reconstructs abnormal data with relatively high accuracy, and this may lead to false positive anomaly detection. In contrast, RCGAN successfully regularizes the generation towards the normal manifold by using $t(\vx)$, with the reconstructed abnormal data resembling digit 0. This result empirically demonstrates that RCGAN offers stronger guarantees for anomaly detection, which is consistent with the theory.}
\label{fig:recon}
\end{figure}

Although previous generative models have the ability to reconstruct normal samples and assign low anomaly score to normal data, these models offer limited guarantees for detecting anomalous samples. To be more specific, these methods rely on the property that the reconstruction $G(E(\vx))$ is close to $\vx$ for normal data and far for anomalous; however, $G(E(\vx))$ does not necessarily yield poor reconstruction (leading to high anomaly scores) for anomalous samples. For example, as shown in \Cref{fig:recon}, an autoencoder trained on normal data can wrongly produce highly accurate reconstructions of abnormal data. Also for previous GAN-based models, it is not guaranteed that the discriminator, trained to distinguish between $\vx \sim q(\vx)$ and ``fake'' samples from the generator, can successfully discriminate between normal and abnormal data. One example is shown in the third row of \Cref{fig:dis} where the discriminators fail to recognize the manifold of normal data.

\section{METHODOLOGY}
\label{sec:meth}
\subsection{Regularize the Discriminator and Generator}
To tackle the limitations mentioned above and enable the GAN to distinguish between normal and abnormal data, we propose a penalty distribution $ t(\vx)$ such that $\vx \sim  t(\vx)$ are considered as adversarial examples for the discriminator during training.

Our method uses ALICE as starting point. Concretely, we propose to include a regularization term $\E_{\vx\sim  t(\vx)}\log (1 - D_{\vx\vz}(\vx, E(\vx)))$ in \Cref{eq:ali}. The first part of our objective function is as follows:
\begin{equation}
\begin{aligned}
\min_{E, G}\max_{D_{\vx\vz}}V_\text{ano}(D_{\vx\vz}, G&, E) = \E_{\vx\sim q(\vx)}[\log D_{\vx\vz}(\vx, E(\vx))]\\
& + \E_{\vz\sim p(\vz)}[\log(1 - D_{\vx\vz}(G(\vz), \vz))]\\
& + \E_{\vx\sim  t(\vx)}[\log (1 - D_{\vx\vz}(\vx, E(\vx)))]
\end{aligned}
\label{eq:aug_xz}
\end{equation}

In our model, the penalty distribution $ t(\vx)$ is chosen to be a random distribution, e.g., Gaussian distribution. These random distributions have broad support so that by adversarial training, the generator and the discriminators are biased towards to the normal manifold as is proved later in this section. As a result, if we consider $G(E(\vx))$ where $\vx$ is anomalous, $G(E(\vx))$ must be close to the normal sample distribution. Consequently, $G(E(\vx))$ must be far from $\vx$, which is the desired outcome for our detection algorithm. This is empirically validated by the examples shown in \Cref{fig:recon}. An RCGAN trained on images of digit 0 successfully biases the generation towards the normal data manifold: the reconstructions of abnormal data resemble digit 0. This leads to large differences between anomalous data and their reconstructions, and consequently a strong guarantee of anomaly detection. These arguments are supported by the theoretical analysis in this section and the superior experimental performance for our model.

Next, we present theoretical results showing that the optimal discriminator and generator distribution are biased more strongly towards normal data. We first derive the optimal discriminator and then the corresponding optimal generator. Consider the following joint distributions:
\begin{itemize}
	\item The encoder joint distribution on normal data $q(\vx, \vz) = q(\vx)e(\vz | \vx)$.
	\item The encoder joint distribution on penalty data $ t(\vx, \vz) =  t(\vx)e(\vz | \vx)$
	\item The generator joint distribution $p(\vx, \vz) = p(\vz)p(\vx | \vz)$
\end{itemize}

The conditional distributions $p(\vx | \vz)$ and $e(\vz | \vx)$ are specified by the generator and the encoder networks respectively. Recall that marginal distributions $q(\vx)$, $ t(\vx)$ and $p(\vz)$ correspond to normal data distribution, penalty distribution and latent variable distribution. The following proposition shows the optimal discriminator $D_{\vx\vz}$:

\begin{proposition}
For fixed generator $G$ and encoder $E$, the optimal discriminator $D_{\vx\vz}^{*}$ from \Cref{eq:aug_xz} is given by:
\begin{equation}
\label{eq:dopt}
\begin{aligned}
D_{\vx\vz}^{*} &= \frac{q(\vx, \vz)}{q(\vx, \vz)+  t(\vx, \vz) + p(\vx, \vz)}\\
&= \frac{q(\vx, \vz)}{(1+ \frac{ t(\vx)}{q(\vx)})q(\vx, \vz) + p(\vx, \vz)}
\end{aligned}
\end{equation}
\end{proposition}

The proof is in the supplementary material (SM). This optimal discriminator considers both normal data distribution and penalty data distribution. This result shows that, unlike classic GANs trained only on normal data, the optimal discriminator in our model is assigning higher probability to more normal data and lower probability to anomalous data with smaller $q(\vx)$. Experiments on synthetic datasets in \cref{sec:syn_exp} further support this conclusion.

Next we will show the optimal generator distribution. Substitute \Cref{eq:dopt} back to \Cref{eq:aug_xz} and let $s(\vx, \vz) = q(\vx, \vz)+  t(\vx, \vz) + p(\vx, \vz)$ and $C(E, G) = V(D^*_{\vx\vz}, G, E)$ for shorthand, it follows that:
\begin{equation}
\label{eq:kl}
\begin{aligned}
C(E, G) &= 2 \; D_{\text{KL}}\Big(\frac{1}{2}
( t(\vx, \vz) + p(\vx, \vz)) \parallel \frac{1}{3}s(\vx, \vz)
\Big) \\
& + D_{\text{KL}} \Big(q(\vx, \vz) \parallel \frac{1}{3}s(\vx, \vz) \Big) - \log \frac{27}{4}
\end{aligned}
\end{equation}
where $D_{\text{KL}}$ denotes Kullback–Leibler divergence.

\begin{theorem}
Given any encoder, the optimal generator distribution $p(\vx, \vz)$ minimizing \Cref{eq:kl} is achieved at

\begin{equation}\label{eq:opt_gen}
p(\vx_i, \vz_j) = \max(0, \beta q(\vx_i, \vz_j) -  t(\vx_i, \vz_j))
\end{equation}

where
\begin{equation}\label{eq:beta}
\beta =
\frac{1 + \sum_{(m, n) \in S_\beta}  t(\vx_m, \vz_n) }
{\sum_{(m, n) \in S_\beta} q(\vx_{m}, \vz_{n}) }
\end{equation}
with $S_\beta = \{ (m,n) \; | \; \beta q(\vx_m, \vz_n) -  t(\vx_m, \vz_n) \ge 0 \}$. \Cref{eq:beta} has a unique solution (note that $\beta$ shows up on both sides of \Cref{eq:beta}), and $1 \le \beta \le 2$. Moreover, $\beta = 1$ whenever $q t = 0$ everywhere (i.e., $q$ and $n$ do not overlap), and $\beta = 2$ whenever $2 q - t \ge 0$ everywhere (e.g., $q=n$).
\end{theorem}

The proof follows Karush-Kuhn-Tucke (KKT) conditions and the convex property of $\beta$. The detailed explanation is given in the SM.

The optimal generator in \Cref{eq:opt_gen} guarantees that anomalous data $\vx$ with low $q(\vx)$ has a poor reconstruction. Since $\beta q(\vx_i, \vz_j) -  t(\vx_i, \vz_j) = (\beta q(\vx_i) -  t(\vx_i))e(\vz_j|\vx_i)$, this theorem indicates that the optimal generator maps the latent variable $\vz$ to $\vx$ for which the normal data probability $q(\vx)$ is high and the penalty distribution $ t(\vx)$ is low. The penalty distribution $t(\vx)$ is used to bias the generator more strongly towards the normal set, removing the ``outliers'' with low $q(\vx)$, as shown in Thm.\ 1 and Prop.\ 1, and therefore improving accuracy overall. In the absence of any information about actual anomalies, this is an effective strategy.

\begin{figure}
\centering
\includegraphics[width=0.8\columnwidth]{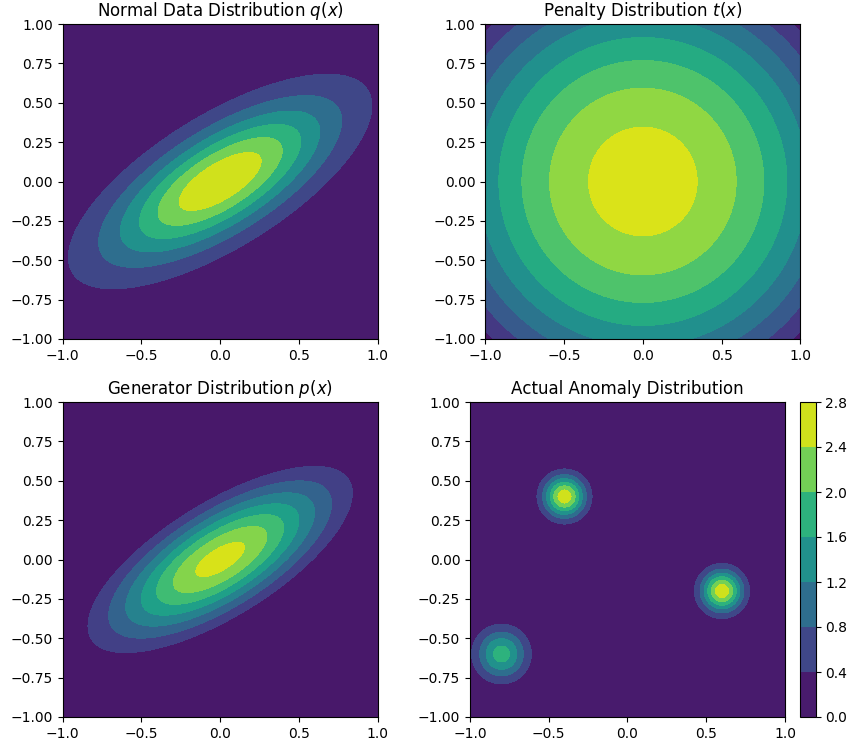}
\caption{An example of normal data distribution $q(\vx)$, penalty distribution $ t(\vx)$, resulting generator distribution in \Cref{eq:opt_gen} and actual anomaly distribution. It shows that our algorithm essentially ``chops off'' regions where $q(\vx)$ is small. This regularization happens independently of how $ t(\vx)$ is related to the actual anomaly distribution.}
\label{fig:gen}
\end{figure}

As an example, assume that $q(\vx)$ has support in a manifold of dimension less than $d$ in a unit cube (assuming as is commonly the case that the input data is normalized). As a result $q(\vx)$ is large inside this manifold and small outside. In contrast, $ t(\vx)$ is chosen to be roughly uniform inside the unit sphere (e.g., a Gaussian distribution with standard deviation 1 and mean 0). In this set up, whenever $q(\vx)$ is large $ t(\vx)$ is small (it doesn't have to be close to zero) and vice versa. In that case our algorithm will provide the correct bias for $G$ and will improve the predictions.  A simple example in $\R^2$ is shown in \Cref{fig:gen}. The normal $q(\vx)$ is a ``narrow'' Gaussian distribution and $t(\vx)$ is roughly uniform in $[-1, 1]^2$. Note the $t(\vx)$ is unrelated to the actual anomalous examples. The resulting generator $p(\vx)$ is computed by minimizing Eq.~(6) using the convex optimization library CVXPY \cite{cvxpy} (for simplicity of presentation, we assume the latent distribution $p(\vz)$ and the conditional distribution $e(\vz|\vx)$ to be constant, so $p(\vx) \sim p(\vx, \vz)$ and $q(\vx) \sim q(\vx, \vz)$). Same as proved in Thm.\ 1, RCGAN correctly biases the generator distribution towards the normal manifold.

\subsection{Generation with Cycle Consistency}
To further guarantee a good reconstruction for normal data $\vx\sim q(\vx)$, we include a second discriminator $D_{\vx\vx}$ as in \cite{li2017alice} to enforce cycle-consistency of $\vx$ and its reconstruction. The cycle-consistency optimization objective function is defined as:
\begin{equation}
\label{eq:ce}
\begin{aligned}
\min_{E, G}\max_{D_{\vx\vx}}V_\text{cycle}(D_{\vx\vx}, G, &E) = \E_{\vx\sim q(\vx)}[\log D_{\vx\vx}(\vx, \vx)]\\
&+ \E_{\vx \sim q(\vx)}[\log(1 - D_{\vx\vx}(\vx, \tilde{\vx}))]
\end{aligned}
\end{equation}
where $\tilde{\vx} = G(E(\vx))$ is the reconstruction of $\vx$. As shown in \cite{li2017alice}, the optimal generator and encoder of the objective in \Cref{eq:ce} leads to $\E_{e(\vz|\vx)}p(\tilde{\vx}|\vz) = \delta(\vx - \tilde{\vx})$, resulting in a perfect reconstruction for $\vx\sim q(\vx)$ theoretically. The optimal discriminator is $D^*_{\vx\vx}(\vx, \tilde{\vx}) = \delta(\vx - \tilde{\vx})$.

The complete minmax optimization objective of the our framework \textit{Regularized Cycle Consistent GAN} (RCGAN) is the sum of \Cref{eq:aug_xz,eq:ce}:
\begin{equation}
\min_{E, G}\max_{D_{\vx\vz}, D_{\vx\vx}} V_\text{ano}(D_{\vx\vz}, G, E) + V_\text{cycle}(D_{\vx\vx}, G, E)
\label{eq:RCGAN}
\end{equation}
After the model is trained on normal data from $p(\vx)$ and adversarial data from $ t(\vx)$ following \Cref{eq:RCGAN}, at the detection phase, the anomaly score assigned to an example $\vx$ is defined as:
\begin{equation}
A(\vx) = 1 - D_{\vx\vx}(\vx, G(E(\vx)))
\label{eq:ano_score}
\end{equation}
The anomaly score $A(\vx)$ describes how well the example $\vx$ is reconstructed, determined by the discriminator $D_{\vx\vx}$. As showed previously, our model enforces a large reconstruction error on anomalous data with low $q(\vx)$ (which is a desirable feature for identification). Meanwhile, the cycle-consistent objective function in \Cref{eq:ce} encourages accurate reconstruction for normal data. This discrepancy endows our model with the ability to discriminate the abnormal from the normal much more reliably. In the next section, numerical experiments on both synthetic and real-world datasets will further demonstrate the effectiveness of our model.

\section{EXPERIMENTS}
\subsection{Towards Better Discriminators}
\label{sec:syn_exp}

\begin{figure*}[htbp]
  \centering
  \includegraphics[width=0.95\textwidth]{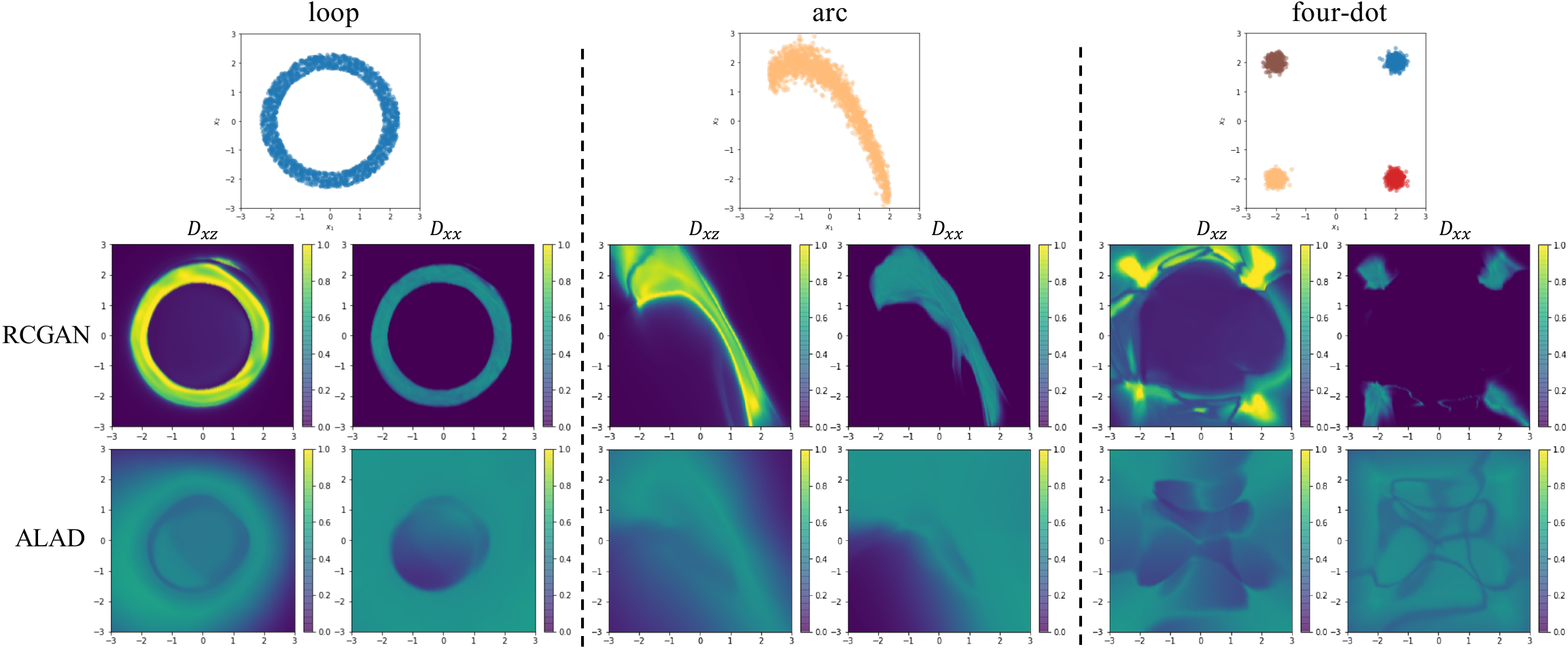}
  \caption{Results on three synthetic datasets: ``loop,'' ``arc'' and ``four-dot.'' The first row shows samples of normal data, and the second and third row show the output probability of discriminators in RCGAN and ALAD respectively. In each dataset, the left column visualizes the output probability of $D_{\vx\vz}$, and the right one shows the output from $D_{\vx\vx}$. These plots show the clear distinction between normal and abnormal sets with RCGAN. ALAD's prediction is much fuzzier.}
  \label{fig:dis}
\end{figure*}

We first test on synthetic dataset in $\R^2 = \{\vx_1, \vx_2\}$. We compare our model with another GAN-based method ALAD. Since both RCGAN and ALAD use discriminators for anomaly detection, we plot the discriminators output after training on normal data. Three cases of normal data distributions $q(\vx)$ are tested on: loop, arc and four-dot. The normal data samples are shown in the first row in \Cref{fig:dis}. The second and the third row contain output by discriminators in RCGAN and ALAD respectively. The output probability given by $D_{\vx\vz}(\vx, E(\vx))$ and $D_{\vx\vx}(\vx, G(E(\vx)))$ over $\vx_1 \in [-3, 3]$, $\vx_2 \in [-3, 3]$ are presented in the first and second column in each case. Higher probability (represented by brighter color) indicates that the discriminator is more confident that $\vx$ is normal. RCGAN and ALAD are trained using the same neural structures and hyper-parameters for fair comparison. We use the multivariate Gaussian $\gN(\vzero, \mI)$ as $ t(\vx)$ in RCGAN.

Discriminators from RCGAN exhibit superior performance, assigning lower normal probability to anomalous data outside the normal data cluster. Notice in the ``four-dot'' case that normal data reside in discontinuous clusters; nevertheless, $D_{\vx\vz}$ and $D_{\vx\vx}$ in RCGAN accurately recognize anomalous data between normal data clusters. This shows that random noise from the penalty distribution $ t(\vx)$, serving as adversarial examples during the training, encourages discriminators to assign low probability to regions where normal examples are missing. This matches with our previous theoretical analysis.

\subsection{Baseline Models}

In our experiments with real-world dataset, we compare RCGAN with following baseline models:

\textbf{Anomaly Detection GAN (ADGAN)} is the first GAN-based anomaly detection model \cite{schlegl2017unsupervised}. After training a DCGAN \cite{radford2015unsupervised} on normal data, test examples are mapped back to corresponding latent variables $\vz$ by minimizing the weighted sum of reconstruction error and feature mapping error via gradient descent.

\textbf{Adversarially Learned Anomaly Detection (ALAD)} in \cite{Zenati2018AdversariallyLA} uses the GAN framework proposed in ALICE and exploit the encoder network to map data back to latent variable space. The anomaly score is the feature mapping error estimated from $D_{\vx\vx}$.

\textbf{Deep Autoencoding Gaussian Mixture Model (DAGMM)} in \cite{zong2018deep} learns an autoencoder for feature extraction and a Gaussian Mixture Model for density estimation. Data with small weighted sum of probabilities predicted by the learned Gaussian mixture are considered anomalous.

\textbf{Deep Structured Energy-Based Model (DSEBM)} trains a network that outputs the energy associated with a data example \cite{zhai2016deep}. Two types of DSEBM, leveraging energy (DSEBM-e) and reconstruction error (DSEBM-r) respectively for anomaly detection, are included for comparison.

\textbf{Deep Support Vector Data Description (DSVDD)} trains a neural network while minimizing the volume of a hypersphere that encloses the network representations of the data \cite{ruff2018deep}. The anomaly score is defined as the Euclidean distance of the data the center of hypersphere.

\textbf{Isolation Forests (IF)} constructs trees by randomly selecting features and then arbitrarily choosing a split value on selected features \cite{liu2008isolation}. The anomaly score of an example is defined as the averaged path length from the root node to the example.

\textbf{One Class Support Vector Machines (OC-SVM)} is a kernel-based method that finds maximum margin hyperplane that separates normal data from the origin \cite{scholkopf2000support}. We use an RBF kernel $K(\vx, \vx')=\exp(-\frac{1}{m} \|\vx - \vx' \|^{2})$ in the experiment, where $m$ is the size of input features.

\textbf{Deep Convolutional Autoencoder (DCAE)} is a classical autoencoder with encoder and decoder based on convolutional neural network \cite{dcae}. The anomaly score is $l_2$ norm of the reconstruction error.

\subsection{Tabular Dataset}

\begin{table*}[!htb]
\caption{Precision, recall and $F_1$ in percent on KDDCUP, Arrhythmia, Thyroid and Musk dataset of RCGAN and benchmark models. The last row is the error bar of RCGAN's performance. The best results for each metric are in bold.}
\centering
\begin{tabular}{c|ccc|ccc|ccc|ccc}
\toprule
\multirow{2}{*}{Model} & \multicolumn{3}{c|}{KDDCUP} & \multicolumn{3}{c|}{Arrhythmia} & \multicolumn{3}{c}{Thyroid} & \multicolumn{3}{c}{Musk}\\
 & Prec. & Recall & $F_{1}$ & Prec. & Recall & $F_{1}$ & Prec. & Recall & $F_{1}$ & Prec. & Recall & $F_{1}$\\
\midrule
IF & 92.16 & 93.73 & 92.94 & 51.47 & 54.69 & 53.03 & 70.13 & 71.43 & 70.27 & 47.96 & 47.72 & 47.51\\
OC-SVM & 74.57 & 85.23 & 79.54 & \textbf{53.97} & 40.82 & 45.18 & 36.39 & 42.39 & 38.87 & - & - & - \\
DSEBM-r & 85.12 & 64.72 & 73.28 & 15.15 & 15.13 & 15.10 & 4.04 & 4.03 & 4.03 & - & - & - \\
DSEBM-e & 86.19 & 64.46 & 73.99 & 46.67 & 45.65 & 46.01 & 13.19 & 13.19 & 13.19 & - & - & -\\
ADGAN & 87.86 & 82.97 & 88.65 & 41.18 & 43.75 & 42.42 & 44.12 & 46.87 & 45.45 & 3.06 & 3.10 & 3.10\\
DAGMM & 92.97 & 94.22 & 93.69 & 49.09 & 50.78 & 49.83 & 47.66 & 48.34 & 47.82 & - & - & -\\
ALAD & 94.27 & 95.77 & 95.01 & 50.00 & 53.13 & 51.52 & 22.92 & 21.57 & 22.22 & 58.16 & 59.03 & 58.37\\
DSVDD & 89.81 & 94.97 & 92.13 & 35.32 & 34.35 & 34.79 & 22.22 & 23.61 & 23.29 & - & - & -\\
\textbf{RCGAN} & \textbf{95.17} & \textbf{96.69} & \textbf{95.92} & 52.73 & \textbf{56.06} & \textbf{54.14} & \textbf{77.08} & \textbf{75.56} &  \textbf{76.26} & \textbf{67.00} & \textbf{66.21} & \textbf{66.49}\\
error bar & 0.28 & 0.29 & 0.28 & 6.6 & 6.8 & 5.8 & 4.3 & 2.7 & 2.8 & 5.06 & 2.53 & 2.62\\
\bottomrule
\end{tabular}
\label{tab:ky}
\end{table*}

\begin{table*}[htbp]
\centering
\caption{Novelty detection on CIFAR-10 dataset by treating each class as normal evaluated by AUROC. Performance with highest mean is in bold.}
\begin{tabular}{c|ccccccc}
\toprule
Normal & DCAE & DSEBM & DAGMM & IF & ADGAN & ALAD & \textbf{RCGAN}\\
\midrule
airplane & 59.1$\pm$5.1 & 41.4$\pm$2.3 & 56.0$\pm$6.9 & 60.1$\pm$0.7 & 67.1$\pm$2.5 & 64.7$\pm$2.6 & \textbf{71.8$\pm$1.5}\\
auto. & 57.4$\pm$2.9 & 57.1$\pm$2.0 & 56.0$\pm$6.9 & 50.8$\pm$0.6 & 54.7$\pm$3.4 & 45.7$\pm$0.8 & \textbf{59.5$\pm$0.7}\\
bird & 48.9$\pm$2.4 & 61.9$\pm$0.1 & 53.8$\pm$4.0 & 49.2$\pm$0.4 & 52.9$\pm$3.0 & \textbf{67.0$\pm$0.7} & 66.2$\pm$0.2\\
cat & 58.4$\pm$1.2 & 50.1$\pm$0.4 & 51.2$\pm$0.8 & 55.1$\pm$0.4 & 54.5$\pm$1.9 & 59.2$\pm$1.1 & \textbf{63.9$\pm$1.7}\\
deer     & 54.0$\pm$1.3 & 73.2$\pm$0.2 & 52.2$\pm$7.3 & 49.8$\pm$0.4 & 65.1$\pm$3.2 & 72.7$\pm$0.6 & \textbf{73.4$\pm$0.9}\\
dog      & \textbf{62.2$\pm$1.8} & 60.5$\pm$0.3 & 49.3$\pm$3.6 & 58.5$\pm$0.4 & 60.3$\pm$2.6 & 52.8$\pm$1.2 & 59.6$\pm$1.1\\
frog     & 51.2$\pm$5.2 & 68.4$\pm$0.3 & 64.9$\pm$1.7 & 42.9$\pm$0.6 & 58.5$\pm$1.4 & 69.5$\pm$1.1 & \textbf{73.0$\pm$1.3}\\
horse    & 58.6$\pm$2.9 & 53.3$\pm$0.7 & 55.3$\pm$0.8 & 55.1$\pm$0.7 & \textbf{62.5$\pm$0.8} & 44.8$\pm$0.4 & 52.5$\pm$0.5\\
ship     & \textbf{76.8$\pm$1.4} & 73.9$\pm$0.3 & 51.9$\pm$2.4 & 74.2$\pm$0.6 & 75.8$\pm$4.1 & 73.4$\pm$0.4 & 73.4$\pm$3.2\\
truck    & \textbf{67.3$\pm$3.0} & 63.6$\pm$3.1 & 54.2$\pm$5.8 & 58.9$\pm$0.7 & 66.5$\pm$2.8 & 43.2$\pm$1.3 & 57.2$\pm$0.6\\
\midrule
mean & 59.4 & 60.3 & 54.4 & 55.5 & 61.8 & 59.3 & \textbf{65.1}\\
\bottomrule
\end{tabular}
\label{tab:cifar10}
\end{table*}

We tested on four tabular datasets: KDDCUP, Arrhythmia, Thyroid and MUSK. Note we choose Thyroid and MUSK dataset for their low anomaly ratio (2.5\% and 3.2\%) to examine RCGAN's robustness in demanding scenario. The experiment setups follow \cite{Zenati2018AdversariallyLA, zong2018deep}:
\begin{itemize}
    \itemsep0em
	\item \textbf{KDDCUP}. The original KDDCUP network intrusion dataset \cite{Dua:2019} contains 494,021 samples with 34 categorical and 7 continuous features. During pre-processing, categorical features are encoded using one-hot representation, and the final data examples have 121 dimensions. Data labelled as ``non-intrusion'' (consisting of 20\% in the dataset) are treated as anomalies since they are in a minority group. In the test phase, the top 20\% of test data with the highest anomaly scores $A(\vx)$ are predicted as anomalies.

	\item \textbf{Arrhythmia}. The Cardiac Arrhythmia dataset \cite{Dua:2019} has 452 instances with 274 attributes, and each instance is classified into one of 16 groups. The smallest classes, including 3, 4, 5, 7, 8, 9, 14, and 15, consist of 15\% of the entire samples and are treated as the anomaly class. The remaining groups are considered as normal data. The top 15\% of the test data with the highest anomaly scores $A(\vx)$ are labeled as anomalies.

	\item \textbf{Thyroid}. The Thyroid disease dataset from \cite{Dua:2019} is a three-class classification dataset with 3,772 instances and 6 continuous attributes. The ``hyperfunction'' class, consisting of 2.5\% of the dataset set, is treated as anomaly. Therefore, the top 2.5\% of the test data with the highest anomaly scores $A(\vx)$ are inferred as abnormal.

	\item \textbf{Musk}. The Musk Anomaly Detection dataset processed by \cite{Rayana2016} is originally a multi-class classification dataset on musk molecular with 3,062 instances and 166 attributes. The musk category 213 and 211 are regarded as anomalous, and the overall anomaly ratio is
	$3.2\%$.
\end{itemize}

We take 80\% of data by random sampling for training and the remaining for test in KDDCUP and Arrhythmia. For Thyroid, 50\% of data are randomly chosen for training, and the rest are for test. For Musk, we follow the original data split. Models are evaluated by precision, recall and F1 scores of the anomaly examples predicted. The results are summarized in \Cref{tab:ky}. We collect performance of benchmark models from \cite{zong2018deep} and \cite{Zenati2018AdversariallyLA}, except that ALAD on Arrhythmia and DSVDD are run by us. Results for RCGAN are averaged over 10 runs. The neural structure for discriminators, generators and encoders used in our models are standard fully connected layers with non-linear gate. We use $\gN(\vzero, \mI)$ as $ t(\vx)$ in RCGAN.

For a clearer comparison, we also provide the error bar for RCGAN's performance in the last row in \cref{tab:ky}. On all four datasets, RCGAN outperforms previous anomaly detection models by significant margins. Especially compared to previous GAN-based methods, the improvement achieved by RCGAN demonstrates the effectiveness of applying a penalty distribution to the adversarial training. On larger datasets, e.g. KDDCUP, Arrhythmia and MUSK, RCGAN's improvements are statistically significant.

\subsection{Image Dataset}
We further test on the image dataset CIFAR-10 as a novelty detection task. Ten distinct datasets are generated by regarding each image category as the normal class. We follow the train/test split in the original dataset. The metrics for evaluation is area under the receiver operating curve (AUROC), averaged on 10 runs. We introduce another type of anomaly score function, proposed in ALAD \cite{Zenati2018AdversariallyLA}:
\begin{equation}
A(\vx) = \|l_{\vx\vx}(\vx, \vx) - l_{\vx\vx}(\vx, G(E(\vx))) \|_2
\label{eq:ano_fm}
\end{equation}
where $l_{\vx\vx}$ denotes the last layer before the logit output in the discriminator $D_{\vx\vx}$. $A(\vx)$ is motivated by the matching loss used to stabilize training of GANs in \cite{salimans2016improved}. Again, our model shows an overall strong performance and achieves highest mean performance in five datasets with statistically significant margins. Notably, RCGAN outperforms baseline models on the average performance across 10 datasets (shown in the last row). Compared with previous GAN-based anomaly detection algorithm, ADGAN and ALAD, RCGAN shows overall competitive results. The hyper-parameters and neural structures of RCGAN closely follow ALAD for a fair comparison. More details on training will be provided in the final version.

\section{DISCUSSION}
\textbf{\quad Ablation Study.}
To further demonstrate the effectiveness of leveraging the penalty distribution $ t(\vx)$, we test RCGAN with and without $ t(\vx)$ in the adversarial training (by removing the last term in \Cref{eq:aug_xz}). Results on Thyroid and Arrhythmia dataset are shown in \Cref{tab:abl}. Utilizing the penalty distribution shows consistent improvement over the model without $ t(\vx)$. This improvement again confirms the effectiveness of our model.

\begin{table}
\caption{Performance of RCGAN with and without leveraging $ t(\vx)$ in the adversarial training.}
\centering
\resizebox{1\columnwidth}{!}{
\begin{tabular}{c|ccc|ccc}
 \toprule
 \multirow{2}{*}{Model} & \multicolumn{3}{c|}{Arrhythmia} & \multicolumn{3}{c}{Thyroid}\\
 & Prec. & Recall & $F_{1}$ & Prec. & Recall & $F_{1}$\\\midrule
 w/o $ t(\vx)$  & 50.00 & 51.93 & 50.85 & 64.06 & 66.35 & 64.94\\
 with $ t(\vx)$ & 52.73 &  56.06 & 54.14 & 77.08 & 75.56 & 75.80\\
\bottomrule
\end{tabular}
}
\label{tab:abl}
\end{table}

\begin{table}
\caption{Using different random distributions $\gN(\vzero, \mI)$, $\gN(\vzero, 2\mI)$ and $\gU(-\vone, \vone)$ as $ t(\vx)$ for unsupervised learning tasks on KDDCUP and Thyroid dataset.}
\centering
\resizebox{1\columnwidth}{!}{
\begin{tabular}{c|ccc|ccc}
\toprule
\multirow{2}{*}{$ t(\vx)$} & \multicolumn{3}{c|}{KDDCUP} & \multicolumn{3}{c}{Thyroid}\\
 & Prec. & Recall & $F_{1}$ & Prec. & Recall & $F_{1}$\\\midrule
 $\gN(\vzero, \mI)$ & 95.17 & 96.69 & 95.92 & 77.08 & 75.56 & 75.80 \\
$\gN(\vzero, 2\mI)$ & 95.26 & 96.78 & 96.01 & 75.83 & 75. 93& 76.48 \\
 $\gU(-\vone, \vone)$ & 93.85 & 95.45 & 94.60 & 75.34 & 77.76 & 76.48\\
\bottomrule
\end{tabular}
}
\label{tab:noise}
\end{table}

\textbf{Choices of the Penalty Distribution.}
In all experiments mentioned previously, the penalty distributions $ t(\vx)$ are chosen to be Gaussian distribution with zeros mean and identity variance. In this subsection, we evaluate the effect of using different types of random distributions for $ t(\vx)$, including two Gaussian distributions $\gN(\vzero, \mI)$ (used for experiments in this paper), $\gN(\vzero, 2\mI)$, and a uniform distribution $\gU(-\vone, \vone)$. Our model is robust and produces consistent performance improvements using anyone of these random distributions, as summarized in \Cref{tab:noise}.

\textbf{Function of the Penalty Distribution.}
The penalty distribution introduced in our model is not designed to ensemble actual anomalous data. Our method still works if $ t(\vx)$ is not close to real anomalies. We still have good reconstruction for data with high $q(\vx)$, i.e., high certainty that $\vx$ is normal, and poor reconstruction for data with low $q(\vx)$ (which is favorable). The conclusion from Thm.\ 1 is unrelated to actual anomalies. In the case that $ t(\vx)$ coincides with normal data $q(\vx)$, we can systematically modify $ t(\vx)$ appropriately to avoid this issue, since we have samples (training data) from $q(\vx)$.

\textbf{Potential Limitation.} A failure case (weakness) of RCGAN corresponds to situations where the following conditions are met: (1) an overwhelming amount of training data is available, and (2) the real anomalies are very far from $q(\vx)$. With a large amount of training data, we should expect the algorithm to recognize that samples drawn from the tail of $q(\vx)$, but not anomalous, are normal. However, the introduction of $t(\vx)$ “chops off” the tail of $q(\vx)$, leading to potentially wrong predictions. RCGAN was designed for cases when the training data is more limited or noisy and some amount of regularization (provided by $t(\vx)$) is required to improve the prediction. 

\section{CONCLUSION}
In this paper, we propose a GAN-based anomaly detection approach which explicitly introduces the penalty distribution in the adversarial learning. Theoretical analysis shows that the introduction of the penalty distribution offers stronger guarantees that our model will correctly distinguish normal from abnormal data. In our numerical experiments, we show that our model consistently outperforms baseline anomaly detection models on four tabular datasets and ten image datasets. We also demonstrated that the performance of the algorithm is relatively insensitive to the choice of the penalty distribution. For future work, we would like to extend the usage of penalty distribution to other generative models besides GAN-based framework.

\ack This research is funded by Ford Motor Company. We also would like to thank anonymous reviewers for their valuable feedback.

\appendix

\section{Proofs}

\subsection{Proof of Proposition 1}
In \Cref{eq:aug_xz}, the discriminator $D_{xz}$ is trained to maximize the quantity $V_\text{ano}(D_{\vx\vz}, G, E)$, which can be rewritten using the three joint distributions above:
\begin{equation}
\label{eq:vged}
\begin{aligned}
&\min_{E, G}\max_{D_{\vx\vz}}V_\text{ano}(D_{\vx\vz}, G, E) = \E_{\vx, \vz \sim q(\vx, \vz)}\log D_{\vx\vz}(\vx, \vz) + \\
& \E_{\vx, \vz \sim p(\vx, \vz)}\log(1 - D_{\vx\vz}(\vx, \vz)) + \E_{\vx, \vz \sim  t(\vx, \vz)}\log (1 - D_{\vx\vz}(\vx, \vz))) \\
& = \int_{\vx, \vz} q(\vx, \vz)\log D_{\vx\vz} + \int_{\vx, \vz} ( t(\vx, \vz) + p(\vx, \vz))\log (1 - D_{\vx\vz})
\end{aligned}
\end{equation}
Recall that the function $a\log(x) + b\log(1 - x)$ achieves its maximum at $x = \frac{a}{a+b}$. Therefore, for fixed generator $G$ and encoder $E$, the optimal discriminator $D_{\vx, \vz}$ is:
\begin{equation}
\begin{aligned}
D_{\vx, \vz}^{*} = \frac{q(\vx, \vz)}{(1+ \frac{ t(\vx)}{q(\vx)})q(\vx, \vz) + p(\vx, \vz)}
\end{aligned}
\end{equation}

\subsection{Proof of Theorem 1}
Substitute \Cref{eq:dopt} back to \Cref{eq:aug_xz} and let $s(\vx, \vz) = q(\vx, \vz)+  t(\vx, \vz) + p(\vx, \vz)$ and $C(E, G) = V(D^*_{\vx\vz}, G, E)$ for shorthand, we have:
\begin{equation}
\label{eq:sm_kl}
\begin{aligned}
C(E, G) & = \int_{\vx, \vz}q(\vx, \vz)\log\frac{q(\vx, \vz)}{\frac{1}{3}s(\vx, \vz)} + 2\log2 - 3\log3\\
&+ 2\int_{\vx, \vz}\frac{ t(\vx, \vz) + p(\vx, \vz)}{2}\log\frac{\frac{1}{2}( t(\vx, \vz) + p(\vx, \vz))}{\frac{1}{3}s(\vx, \vz)}\\
& = 2 \; D_{\text{KL}}(\frac{ t(\vx, \vz) + p(\vx, \vz)}{2}\|\frac{s(\vx, \vz)}{3}) \\
&+ D_{\text{KL}}(q(\vx, \vz)\|\frac{s(\vx, \vz)}{3}) + 2 \log 2 - 3 \log 3
\end{aligned}
\end{equation}
where $s(\vx, \vz) =  t(\vx, \vz) + p(\vx, \vz) + q(\vx, \vz)$. Consider the discrete case, and let $p_{ij}$, $q_{ij}$ and $t_{ij}$ denote $p_{\vx_i, \vz_j}$, $q_{\vx_i, \vz_j}$ and $t_{\vx_i, \vz_j}$. Given fixed encoder network, the optimization problem in \Cref{eq:sm_kl} can be rewritten as:
\begin{equation}
\begin{aligned}
&\underset{p_{ij}}{\text{min}} \sum_{i, j}(t_{ij} + p_{ij})\log(t_{ij} + p_{ij}) - \sum_{i, j}q_{ij}\log(t_{ij} + p_{ij} + q_{ij}) \\
&- \sum_{i, j}(t_{ij} + p_{ij})\log(t_{ij} + p_{ij} + q_{ij})\\
&\ \text{s.t. \; } \  p_{ij} \ge 0, \;\; \sum_{i, j} p_{ij} = 1
\end{aligned}
\end{equation}
The Lagrangian is:
\begin{equation}
\label{eq:sm_lag}
\begin{aligned}
&\gL = \lambda \; (\sum_{i, j}p_{ij} - 1) - \sum_{i, j}\mu_{ij}p_{ij} + \sum_{i, j}(t_{ij} + p_{ij})\log(t_{ij} + p_{ij})\\
&- \sum_{i, j}(t_{ij} + p_{ij})\log(t_{ij} + p_{ij} + q_{ij}) - \sum_{i, j}q_{ij}\log(t_{ij} + p_{ij} + q_{ij})
\end{aligned}
\end{equation}
where $\lambda$ and \{$\mu_{ij}$\} are Karush–Kuhn–Tucker (KKT) multipliers. Following KKT conditions, we have:
\begin{equation}
\label{eq:kkt_cond}
\frac{\partial \gL}{\partial p_{ij}} = 0, \quad
\mu_{ij}p_{ij} = 0, \quad
\mu_{ij} \geq 0
\end{equation}
For all $i, j$, the first condition in \Cref{eq:kkt_cond} leads to:
\begin{equation}
\label{eq:sm_par}
\log\frac{ t_{ij} + p_{ij}}{ t_{ij} + p_{ij} + q_{ij}} + \lambda - \mu_{ij} = 0
\end{equation}
If $p_{ij} \neq 0$, then $\mu_{ij} = 0$ and it becomes:
\begin{equation}
p_{ij} = \beta q_{ij} -  t_{ij}
\end{equation}
where $\beta = \frac{1}{e^{\lambda} - 1}$. Recall $\sum_{i, j}p_{ij} = 1$, then $\beta = \frac{1 + \sum{t_{mn}}}{\sum{q_{mn}}}$ for $p_{mn}\neq 0$.

Next, we will derive the condition for $p_{ij} = 0$. If $p_{ij} = 0$, from \Cref{eq:sm_par} we have:
\begin{equation}
\begin{aligned}
\mu_{ij} = \log & \frac{ t_{ij}}{ t_{ij} + q_{ij}} + \lambda \geq 0, \quad q_{ij} \leq (e^{\lambda} - 1) t_{ij}, \quad
q_{ij} \leq \frac{1}{\beta}  t_{ij}
\end{aligned}
\end{equation}
The inequality in the first line is from the KKT conditions. In sum, in the discrete case, the optimal generator distribution given any encoder is:
\begin{equation}
p(\vx_i, \vz_j) = \max(0, \beta q(\vx_i, \vz_j) -  t(\vx_i, \vz_j))
\end{equation}
where 
\begin{equation}
\beta = 
\frac{1 + \sum_{(m, n) \in S_\beta}  t(\vx_m, \vz_n) }
{\sum_{(m, n) \in S_\beta} q(\vx_{m}, \vz_{n}) }
\end{equation}

with $S_\beta = \{ (m,n) \; | \; \beta q(\vx_m, \vz_n) -  t(\vx_m, \vz_n) \ge 0 \}$. We now prove a few additional properties of the solution. The set $S_1$ must be non-empty. For $\beta \ge 1$, the set $S_\beta$ increases monotonously and therefore cannot be empty. Denote
\[ f(\beta) = 
\frac{1 + \sum_{(m, n) \in S_\beta}  t(\vx_m, \vz_n) }
{\sum_{(m, n) \in S_\beta} q(\vx_{m}, \vz_{n}) }
\]
Since
\begin{equation}\label{eq:one}
\sum_{(m, n) \in S_\beta}  t(\vx_m, \vz_n) \ge 0, \qquad
\sum_{(m, n) \in S_\beta} q(\vx_{m}, \vz_{n}) \le 1
\end{equation}
So $f(\beta) \ge 1$. All solutions $\beta = f(\beta)$ therefore satisfy $\beta \ge 1$.

Denote $S^{\beta,\text{c}}$ the complement of the set $S^\beta$. Then:
\begin{align*} 
f(\beta) & = \frac{1 + \sum_{(m, n) \in S^\beta}  t(\vx_m, \vz_n) }
{\sum_{(m, n) \in S^\beta} q(\vx_{m}, \vz_{n}) } = \frac{2 - \sum_{(m, n) \in S^{\beta,\text{c}}}  t(\vx_m, \vz_n) }
{\sum_{(m, n) \in S^\beta} q(\vx_{m}, \vz_{n}) } \\
& \le \frac{2 - \beta \sum_{(m, n) \in S^{\beta,\text{c}}} q(\vx_m, \vz_n) }
{\sum_{(m, n) \in S^\beta} q(\vx_{m}, \vz_{n}) } = \beta + \frac{2 - \beta}{\sum_{(m, n) \in S^\beta} q(\vx_{m}, \vz_{n}) }
\end{align*}
Consequently, $f(2) \le 2$, and all solutions $\beta = f(\beta)$ are such that $\beta \le 2$.

The function
\[ S_p(\beta) = \sum_{mn} 
\max(0, \beta q(\vx_m,\vz_n) -  t(\vx_m,\vz_n))
\]
is a continuous convex and monotonically increasing function of $\beta$ (although not differentiable). But we have:
\[ 
S_p(\beta) = 1 + (\beta - f(\beta)) \sum_{(m, n) \in S^\beta} q(\vx_{m}, \vz_{n})
\]
Since $S_p(1) \le 1$, $S_p(2) \ge 1$, and $S_p$ continuous convex and strictly increasing, then there exists a unique $\beta$ ($1 \le \beta \le 2$) for which $S_p(\beta) = 1$. Therefore, the equation $\beta = f(\beta)$ has a unique solution, and $1 \le \beta \le 2$.

If the solution $\beta = 2$, then from the definition of $f$ we see that $S^{\beta,\text{c}}$ must be empty. So $2 q - t \ge 0$ everywhere. The converse is true. If $\beta = 1$, then $q \neq 0$ implies $t = 0$ (so that $f(1)$ can be equal to 1; see \Cref{eq:one}). For all indices, either $q$ or $t$ is 0 at the corresponding index. This can be written as $q t = 0$ everywhere, or equivalently the support of $q$ and $t$ do not overlap.

\bibliography{ecai}

\end{document}